\def\year{2021}\relax
\documentclass[letterpaper]{article} 
\usepackage{aaai21}  
\usepackage{times}  
\usepackage{helvet} 
\usepackage{courier}  
\usepackage[hyphens]{url}  
\usepackage{graphicx} 
\usepackage{natbib}  
\usepackage{caption} 
\usepackage{amsfonts,xcolor,multicol}
\usepackage[tbtags]{amsmath}
\usepackage[framemethod=TikZ]{mdframed}
\usepackage{textcomp}
\usepackage{subcaption}
\usepackage{algorithm,algpseudocode}
\usepackage{stfloats}
\usepackage{booktabs}
\usepackage{multirow}
\usepackage{stackengine}
\stackMath

\DeclareMathOperator*{\argmax}{arg\,max}
\newcommand{\ours}{\textsc{HAL}}
\newcommand{\dataset}{\mathcal{D}}
\newcommand{\epsmem}{\mathcal{M}}
\newcommand{\anc}{e}

\newcommand\numberthis{\addtocounter{equation}{1}\tag{\theequation}}
\newcommand{\SKIP}[1]{}

\def\forkindep{\mathrel{\raise0.2ex\hbox{\ooalign{\hidewidth$\vert$\hidewidth\cr\raise-0.9ex\hbox{$\smile$}}}}}
\renewcommand{\eqref}[1]{Eq.~\ref{#1}}
\title{Using Hindsight to Anchor Past Knowledge in Continual Learning}
\author {
    Arslan Chaudhry$^{1,}$\footnote{Correspondence to arslan.chaudhry@eng.ox.ac.uk.}, 
    Albert Gordo$^2$, 
    Puneet K. Dokania$^1$,
    Philip Torr$^1$,
    David Lopez-Paz$^2$ \\
}
\affiliations {
    $^1$University of Oxford, $^2$Facebook AI
}

\begin{document}

\maketitle

\begin{abstract}
In continual learning, the learner faces a stream of data whose distribution changes over time. Modern neural networks are known to suffer under this setting, as they quickly forget previously acquired knowledge.
To address such catastrophic forgetting, many continual learning methods implement different types of \emph{experience replay}, re-learning on past data stored in a small buffer known as episodic memory.
In this work, we complement experience replay with a new objective that we call ``anchoring'', where the learner uses bilevel optimization to
update its knowledge on the current task, while keeping intact predictions on some \emph{anchor points} of past tasks.
These anchor points are learned using gradient-based optimization to maximize forgetting, which is approximated by fine-tuning the currently trained model on the episodic memory of past tasks. 
Experiments on several supervised learning benchmarks for continual learning demonstrate that our approach improves the standard experience replay in terms of both accuracy and forgetting metrics and for various sizes of episodic memory. 
\end{abstract}

\section{Introduction} \label{sec:intro}

We study the problem of \emph{continual learning}, where a machine learning model experiences a sequence of tasks.
Each of these tasks is presented as a stream of input-output pairs, where each pair is drawn identically and independently (iid) from the corresponding task probability distribution.
Since the length of the learning experience is not specified a priori, the learner can only assume a \emph{single pass over the data} and, due to space constraints, store nothing but a few examples in a small \emph{episodic memory}.
At all times during the lifetime of the model, predictions on examples from any task may be requested.
Addressing continual learning is an important research problem, since it would enable the community to move past the assumption of ``identically and independently distributed data'', and allow a better deployment of machine learning in-the-wild.
However, continual learning presents one major challenge, \emph{catastrophic forgetting}~\citep{mccloskey1989catastrophic}.
That is, as the learner experiences new tasks, it quickly forgets previously acquired knowledge.
This is a hindrance especially for state-of-the-art deep learning models, where all parameters are updated after observing each example.

Continual learning has received increasing attention from the scientific community during the last decade.
The state of the art algorithms for continual learning fall into three categories.
First, {\em regularization-based} approaches reduce forgetting by restricting the updates in model parameters that were important for previous tasks~\citep{Kirkpatrick2016EWC,Rebuffi16icarl,aljundi2017memory,chaudhry2018riemannian,nguyen2017variational}. However, when the number of tasks are large, the regularization of past tasks becomes obsolete, leading to the representation drift~\citep{titsias2019functional}.
Second, {\em modular approaches}~\citep{rusu2016progressive,lee2017lifelong} add new modules to the learner as new tasks are learned. While modular architectures overcome forgetting by design, the memory complexity of these approaches scales with the number of tasks.
Third, the {\em memory-based} methods~\citep{lopez2017gradient,hayes2018memory,isele2018selective,riemer2018learning,chaudhry2019agem} store a few examples from past tasks in an ``episodic memory'', to be revisited when training for a new task.
Contrary to modular approaches, memory-based methods add a very small memory overhead for each new task. 
Memory-based methods are the reigning state-of-the-art, but their performance remains a far cry from a simple oracle accessing all the data at once, hence turning the continual learning experience back into a normal supervised learning task.
Despite intense research efforts, such gap in performance renders the problem of continual learning an open research question.

\paragraph{Contribution}
We propose Hindsight Anchor Learning (\ours{}), a continual learning approach to improve the performance of memory-based continual learning algorithms.
\ours{} leverages bilevel optimization to regularize the training objective with one representational point per class per task, called \emph{anchors}. 
%
These anchors are constructed via gradient ascent in the image space, by maximizing one approximation to the forgetting loss for the current task throughout the entire continual learning experience. 
We estimate the amount of forgetting that the learner would suffer on these anchors if it were to be trained on future tasks in \emph{hindsight}: that is, by measuring forgetting on a temporary predictor that has been fine-tuned on the episodic memory of past tasks.
Anchors learned in such a way lie close to the classifier's decision boundary, as visualized in Figure~\ref{fig:mnist_visualize}.
Since points near the decision boundary are the easiest to forget when updating the learner on future tasks, keeping prediction invariant on such anchors preserves the performance of previous tasks effectively.
In sum, the overall parameter update of \ours{} uses nested optimization to minimize the loss of the current mini-batch, while keeping the predictions of all anchors invariant.

\paragraph{Results}
We compare \ours{} to EWC \citep{Kirkpatrick2016EWC}, ICARL~\citep{Rebuffi16icarl}, VCL~\citep{nguyen2017variational}, AGEM \citep{chaudhry2019agem}, experience replay \citep{hayes2018memory, riemer2018learning}, MER \citep{riemer2018learning}, and MIR \citep{mir_aljundi} across four commonly used benchmarks in supervised continual learning (MNIST permutations, MNIST rotations, split CIFAR-100, and split miniImageNet).
In these experiments, \ours{} achieves state-of-the-art performance, improving accuracy by up to 7.5\% and reducing forgetting by almost 23\% over the experience replay baseline.
We show that these results hold for various sizes of episodic memory (between $1$ and $5$ examples per class per task).

We now begin our exposition by reviewing the continual learning setup.
The rest of the manuscript then presents our new algorithm \ours{} (Section~\ref{sec:method}), showcases its empirical performance (Section~\ref{sec:experiments}), surveys the related literature (Section~\ref{sec:related_work}), and offers some concluding remarks (Section~\ref{sec:conclusion}). 

\section{Continual learning setup} \label{sec:setup}

In continual learning, a learner experiences a stream of data triplets $(x_i, y_i, t_i)$ containing an input $x_i$, a target $y_i$, and a task identifier $t_i \in \mathcal{T} = \{1, \ldots, T\}$.
Each input-target pair $(x_i, y_i) \in \mathcal{X} \times \mathcal{Y}_{t_i}$ is an identical and independently distributed example drawn from some unknown distribution $P_{t_i}(X, Y)$, representing the $t_i$-th learning task.
We assume that the tasks are experienced in order ($t_i \leq t_j$ for all $i \leq j$), and that the total number of tasks $T$ is not known a priori.
Under this setup, our goal is to estimate a predictor $f_{\theta} = (w \circ \phi) : \mathcal{X} \times \mathcal{T} \to \mathcal{Y}$, parameterized by $\theta \in \mathbb{R}^P$, and composed of a feature extractor $\phi : \mathcal{X} \to \mathcal{H}$ and a classifier $w : \mathcal{H} \to \mathcal{Y}$, that minimizes the multi-task error
\begin{equation}
    \frac{1}{T} \sum_{t=1}^T \mathbb{E}_{(x, y) \sim P_{t}}\left[\, \ell(f(x, t), y) \,\right],
    \label{eq:multitask}
\end{equation}
where $\mathcal{Y} = \cup_{t \in \mathcal{T}} \mathcal{Y}_{t}$, and $\ell : \mathcal{Y} \times \mathcal{Y} \to \mathbb{R}$ is a loss function.

Inspired by prior literature in continual learning, \citep{lopez2017gradient,hayes2018memory,riemer2018learning,chaudhry2019agem}, we consider streams of data that are \emph{experienced only once}.
Therefore, the learner cannot revisit any but a small number of data triplets chosen to be stored in a small episodic memory $\mathcal{M}$.
More specifically, we consider tiny ``ring'' episodic memories, which contain the last $m$ observed examples per class for each of the experienced tasks, where $m \in \{1, 3, 5\}$.
That is, considering as variables the number of experienced tasks $t$ and examples $n$, we study continual learning algorithms with a $O(t)$ memory footprint.

Following~\citet{lopez2017gradient} and \citet{chaudhry2018riemannian}, we monitor two statistics to evaluate the quality of continual learning algorithms: \emph{final average accuracy}, and \emph{final maximum forgetting}.
First, the final average accuracy of a predictor is defined as
\begin{equation}
    \text{Accuracy} = \frac{1}{T} \sum_{j=1}^T a_{T, j},
    \label{eq:accuracy}
\end{equation}
where $a_{i,j}$ denotes the test accuracy on task $j$ after the model has finished experiencing task $i$.
That is, the final average accuracy measures the test performance of the model at every task after the continual learning experience has finished.
Second, the final maximum forgetting is defined as
\begin{equation}
    \text{Forgetting} = \frac{1}{T-1} \sum_{j=1}^{T-1} \max_{l \in \{1, \ldots, T-1\}} (a_{l, j} - a_{T, j}),
    \label{eq:forgetting}
\end{equation}
that is, the decrease in performance for each of the tasks between their peak accuracy and their accuracy after the continual learning experience has finished.

Finally, following~\citet{chaudhry2019agem}, we use the first $k < T$ tasks to cross-validate the hyper-parameters of each of the considered continual learning algorithms.
These first $k$ tasks are not considered when computing the final average accuracy and maximum forgetting metrics.

\section{Hindsight Anchor Learning (\ours{})}
\label{sec:method}

The current state of the art algorithms for continual learning are based on experience replay~\citep{hayes2018memory,riemer2018learning,chaudhry2019er}.
These methods update the model $f_\theta$ while storing a small amount of past observed triplets in an episodic memory $\mathcal{M} = \{(x', y', t')\}$. For a new minibatch of observations $\mathcal{B} := \{(x, y, t)\}$ from task $t$, the learner samples a mini-batch $\mathcal{B}_\mathcal{M}$ from $\mathcal{M}$ at random, and employ the rule 
$\theta \leftarrow \theta - \alpha \cdot \nabla_\theta \, \ell(\mathcal{B} \cup \mathcal{B}_\mathcal{M})$ to update its parameters, where
\begin{align}
    \ell(\mathcal{A}) = \frac{1}{|\mathcal{A}|} \sum_{(x, y, t) \in \mathcal{A}} \ell(f_\theta(x,t), y)
\end{align}
denotes the average loss across a collection of triplets $\mathcal{A}~=~\mathcal{B} \cup \mathcal{B}_\mathcal{M}$. In general, $\mathcal{B}_\mathcal{M}$ is constructed to have the same size as  $\mathcal{B}$, but it can be smaller if the episodic memory $\mathcal{M}$ does not yet contain enough samples.

The episodic memory $\mathcal{M}$ reminds the predictor about how to perform past tasks using only a small amount of data.
As such, the behaviour of the predictor on past tasks outside the data stored in $\mathcal{M}$ is not guaranteed.
Moreover, since $\mathcal{M}$ is usually very small, the performance of the predictor becomes sensitive to the choice of samples stored in the episodic memory.
Because of this reason, we propose to further fix the behaviour of the predictor at a collection of carefully constructed \emph{anchor points} $e_{t'}$, one per class per  past task $t'$, at each parameter update.

Let us assume that the anchor points $e_{t'}$ are given ---we will see later how to construct them in practice. To constrain the change of the predictor at these anchor points, we propose a two-step parameter update rule: 
\begin{align}
    & \tilde{\theta} \leftarrow \theta - \alpha \nabla_\theta \, \ell(\mathcal{B} \cup \mathcal{B}_\mathcal{M}), \nonumber \\
    & \text{\footnotesize$\theta \leftarrow \theta - \alpha \nabla_\theta \biggl( \ell(\mathcal{B} \cup \mathcal{B}_\mathcal{M}) + \lambda \sum_{t' < t} \left(f_\theta(e_{t'}, t') - f_{\tilde{\theta}}(e_{t'}, t')\right)^2 \biggr).$}
    \label{eq:main_update}
\end{align}

The first step computes a temporary parameter vector $\tilde{\theta}$ by minimizing the loss at a minibatch from the current task $t$, and the episodic memory of past tasks (this is the usual experience replay parameter update).
The second step employs a nested optimization to perform the actual update of the parameter $\theta$, which trades-off the minimization of ($a$) the loss value at the current minibatch and the episodic memory, as well as ($b$) changes in predictions at the anchor points for all past tasks.
The proposed rule not only updates the predictor conservatively, thereby reducing forgetting, but also, as shown analytically in Appendix~\ref{sec:approx_grad}, improves the forward transfer by maximizing the inner product between the gradients on $\mathcal{B} \cup \mathcal{B}_\mathcal{M}$ and anchor points.
In this respect, it bears similarity to gradient-based meta-learning approaches~\citep{finn2017model,metareptile,riemer2018learning}.

Next, let us discuss how to choose the anchor points, $e_{t}$ (one per class per task) as to  preserve the performance of the current task throughout the entire learning experience.
Ideally, the anchor points should attempt to minimize the forgetting on the current task as the learner is updated with future tasks.
One could achieve this by letting $e_{t}$ to be an example from the task $t$ that would undergo maximum forgetting during the {\em entire} continual learning experience.
Then, requiring the predictions to remain invariant at $e_{t}$, by using~\eqref{eq:main_update}, could effectively reduce forgetting on the current task.
Mathematically, the desirable $e_t$ for the label $y_t$ is obtained by maximizing the following \emph{Forgetting loss}:
\begin{equation} 
    (e_{t}, y_{t}) \leftarrow \argmax_{(x, y) \sim P_t} \, \ell(f_{\theta_T}(x, t), y_t) - \ell(f_{\theta_t}(x, t), y_t),
    \label{eq:ideal_anchor}
\end{equation}
where $\theta_t$ is the parameter vector obtained after training on task $t$ and $\theta_T$ is the final parameter vector obtained after the entire learning experience.
Thus, keeping the predictions intact on the pair $(e_t, y_t)$ above can effectively preserve the performance of task $t$.
However, the idealistic \eqref{eq:ideal_anchor} requires access to ($a$) the entire distribution $P_t$ to compute the maximization, and ($b$) access to all future distributions $t' > t$ to compute the final parameter vector $\theta_T$. 
Both are unrealistic assumptions under the continual learning setup described in Section~\ref{sec:setup}, as the former requires storing the entire dataset of task $t$, and the latter needs access to future tasks. 

To circumvent ($a$), we can recast \eqref{eq:ideal_anchor} as an optimization problem and {\em learn} the desired $e_t$ by initializing it at random and using $k$ gradient ascent updates for a \emph{given label} $y_t$ in the image space ($\mathcal{X} \in \mathbb{R}^D$). The proposed optimization objective is given by: 
\begin{equation}
    \resizebox{1.0\hsize}{!}{%
    $\max_{e_t \in \mathbb{R}^{D}} \biggl( \underbrace{\ell(f_{\theta_T}(e_t, t), y_t) - \ell(f_{\theta_t}(e_t, t), y_t)}_\text{Forgetting loss} 
    - \gamma \underbrace{(\phi(e_t) - \phi_t)^2}_\text{Mean embedding loss} \biggr),$}    
    \label{eq:anchors_1}
\end{equation}
where the regularizer, given by the mean embedding loss, constrains the search space by trying to push the anchor point embedding towards the mean data embedding. 
We recall that $\phi$ denotes the feature extractor of the predictor, and $\phi_t$ is the neural mean embedding \citep{smola2007hilbert} of all observed examples from task $t$. 
Since the feature extractor is updated after experiencing each data point, the mean embedding $\phi_t$ are computed as running averages.
That is, after observing a minibatch $\mathcal{B} = \{(x, y, t)\}$ of task $t$, we update:
\begin{equation}
    \phi_t \leftarrow \beta \cdot \phi_t + (1 - \beta) \frac{1}{|\mathcal{B}|}\sum_{x \in \mathcal{B}} \phi(x),
    \label{eq:mean_embedding}
\end{equation}
where $\phi_t$ is initialized to zero at the beginning of the learning experience.
In our experiments, we learn one $e_t$ per class for each task. 
We fix the $y_t$ to the corresponding class label, and discard $\phi_t$ after training on task $t$.
Learning $e_t$ in this manner circumvents the requirement of storing the entire distribution $P_t$ for the current task $t$.

Still, ~\eqref{eq:anchors_1} requires the parameter vector $\theta_T$, to be obtained in the distant future after all learning tasks have been experienced. 
To waive this impossible requirement, we propose to \emph{approximate the future by simulating the past}.
That is, instead of measuring the forgetting that would happen after the model is trained for future tasks, we measure the forgetting that happens when the model is fine-tuned for past tasks.
In this way, we say that forgetting is estimated in \emph{hindsight}, using past experiences.
More concretely, after training on task $t$ and obtaining the parameter vector $\theta_t$, we minimize the loss during one epoch on the episodic memory $\mathcal{M}$ to obtain a temporary parameter vector $\theta_{\mathcal{M}}$ that approximates $\theta_T$, and update $e_t$ as:
\vspace{-0.2cm}
\begin{equation}
    e_t \leftarrow e_t + \alpha \nabla_{e_t} \biggl( \ell(f_{\theta_\mathcal{M}}(e_t, t), y_t) - \stackunder[5pt]{{}\ell(f_{\theta_t}(e_t, t), y_t)}{ {}- \gamma (\phi(e_t) - \phi_t)^2 \biggr).}
    \label{eq:anchors_2}
\end{equation}
This completes the description of our proposed algorithm for continual learning, which combines experience replay with anchors learned in hindsight.
We call our approach Hindsight Anchor Learning (\ours{}) and summarize the entire learning process as follows:
\begin{mdframed}[backgroundcolor=yellow!8]
\small
\textbf{Hindsight Anchor Learning (\ours{})}
\begin{itemize}
    \item Initialize $\theta \sim P(\theta)$ and $\{e_t \sim P(e)\}_{t=1}^T$ from normal distributions $P(\theta)$ and $P(e)$, $\mathcal{M} = \{\}$.
    \item Initialize $\mathcal{M} = \{\}$
    \item For each task $t = 1, \ldots, T$:
    \begin{itemize}
        \item For each minibatch $\mathcal{B}$ from task $t$:
        \begin{itemize}
            \item Sample $\mathcal{B}_\mathcal{M}$ from $\mathcal{M}$
            \item Update $\theta$ using \eqref{eq:main_update}
            \item Update $\phi_t$ using \eqref{eq:mean_embedding}
            \item Update $\mathcal{M}$ by adding $\mathcal{B}$ in a FIFO ring buffer  
        \end{itemize}
        \item Fine-tune on $\mathcal{M}$ to obtain $\theta_{\mathcal{M}}$ 
        \item Build $e_t$ using \eqref{eq:anchors_2} $k$ times
        \item Discard $\phi_t$
    \end{itemize}
    \item Return $\theta$.
\end{itemize}
\end{mdframed}

We contrast \ours{} with another idea, Maximally Interfered Retrieval (MIR)~\citep{mir_aljundi} in that MIR \emph{selects} a minibatch from an already populated replay buffer at each training step, whereas \ours{} \emph{writes} to the replay buffer at the end of each task and samples randomly from the replay buffer. 
Furthermore, MIR selects the minibatch by measuring a \emph{one-step} increase in loss incurred by the minibatch whereas \ours{} uses an approximate forgetting loss \eqref{eq:anchors_2} that measures an increase over multiple training steps. 

\section{Experiments} \label{sec:experiments}

We now report experiments on standard image classification benchmarks for continual learning.

\subsection{Datasets and tasks} \label{sub:datasets_and_tasks}
We perform experiments on four supervised classification benchmarks for continual learning. 
\textbf{Permuted MNIST} is a variant of the MNIST dataset of handwritten digits~\citep{lecun1998mnist} where each task applies a fixed random pixel permutation to the original dataset.
This benchmark contains $23$ tasks, each with $1000$ samples from $10$ different classes.
\textbf{Rotated MNIST} is another variant of MNIST, where each task applies a fixed random image rotation (between $0$ and $180$ degrees) to the original dataset.
This benchmark contains $23$ tasks, each with $1000$ samples from $10$ different classes.
\textbf{Split CIFAR} is a variant of the CIFAR-100 dataset~\citep{krizhevsky2009learning,Zenke2017Continual}, where each task contains the data pertaining $5$ random classes (without replacement) out of the total $100$ classes.
This benchmark contains $20$ tasks, each with $250$ samples per each of the $5$ classes.
\textbf{Split miniImageNet} is a variant of the ImageNet dataset~\citep{russakovsky15ImageNet,vinyals2016matching}, containing a subset of images and classes from the original dataset.
This benchmark contains $20$ tasks, each with $250$ samples per each of the $5$ classes.

For all datasets, the first $3$ tasks are used for hyper-parameter optimization (grids available in Appendix~\ref{sec:supp_hyperparam}).
The learner can perform multiple epochs on these three initial tasks that are later discarded for evaluation.

\subsection{Baselines} \label{sub:baselines}
We compare our proposed model \emph{\ours{}} to the following baselines.
\textbf{Finetune} is a single model trained on a stream of data, without any regularization or episodic memory.
\textbf{ICARL} \citep{Rebuffi16icarl} uses nearest-mean-of-exemplar rule for classification and avoids catastrophic forgetting by regularizing over the feature representations of previous tasks using knowledge distillation loss~\citep{hinton2015distilling}. 
\textbf{EWC} \citep{Kirkpatrick2016EWC} is a continual learning method that limits changes to parameters critical to past tasks, as measured by the Fisher information matrix. 
\textbf{VCL} \citep{nguyen2017variational} is a continual learning method that uses online variational inference for approximating the posterior distribution, which is then used to regularize the model. 
\textbf{AGEM} \citep{chaudhry2019agem} is a continual learning method improving on \citep{lopez2017gradient}, which uses an episodic memory of parameter gradients to limit forgetting.
\textbf{MER} \citep{riemer2018learning} is a continual learning method that combines episodic memory with meta-learning to limit forgetting.
\textbf{ER-Ring} \citep{chaudhry2019er} is a continual learning method that uses a ring buffer as episodic memory.
\textbf{MIR} \citep{mir_aljundi} is a continual learning method based on experience replay that selects a minibatch from the episodic memory that incurs the maximum change in loss.
\textbf{Multitask} is an oracle baseline that has access to all data to optimize~\eqref{eq:multitask}, useful to estimate an upper bound on the obtainable accuracy~(\eqref{eq:accuracy}).
\textbf{Clone-and-finetune} is an oracle baseline training one independent model per task, where the model for task $t'$ is initialized by cloning the parameters of the model for task $t'-1$.

All baselines use the same neural network architecture: a perceptron with two hidden layers of 256 ReLU neurons in the MNIST experiments, and a ResNet18, with three times less feature maps across all layers, similar to \citet{lopez2017gradient}, in CIFAR and ImageNet experiments.
The task identifiers are used to select the output head in the CIFAR and ImageNet experiments, while ignored in the MNIST experiments.
Batch size is set to $10$ for both the stream of data and episodic memory across experiments and models.
The size of episodic memory is set between $1$ and $5$ examples per class per task.
The results of VCL are complied by running the official implementation\footnote{\url{https://github.com/nvcuong/variational-continual-learning}}, that only works for fully-connected networks, in our continual learning setup. 
All the other baselines use our unified code base which is available here \footnote{\url{https://github.com/arslan-chaudhry/HindsightAnchor}}.
All experiments are averaged over five runs using different random seeds, where each seed corresponds to a different dataset ordering among tasks.

\subsection{Results}
\label{sub:results}

\begin{table*}[t]
\begin{center}
\begin{small}
\begin{sc}
\caption{Accuracy~(\eqref{eq:accuracy}) and Forgetting~(\eqref{eq:forgetting}) results of continual learning experiments.
Averages and standard deviations are computed over five runs using different random seeds.
When used, episodic memories contain up to one example per class per task.
Last two rows are oracle baselines.}

\label{tab:mnist_comp}
\begin{tabular}{lcccc}
\toprule
\multicolumn{1}{l}{\textbf{Method}} &\multicolumn{2}{c}{\textbf{Permuted MNIST}} &\multicolumn{2}{c}{\textbf{Rotated MNIST}}   \\
\midrule
& Accuracy & Forgetting & Accuracy & Forgetting \\
\midrule
Finetune                             & 53.5 (\textpm 1.46) & 0.29 (\textpm 0.01) & 41.9 (\textpm 1.37) & 0.50 (\textpm 0.01) \\
EWC~\citep{Kirkpatrick2016EWC}       & 63.1 (\textpm 1.40) & 0.18 (\textpm 0.01) & 44.1 (\textpm 0.99) & 0.47 (\textpm 0.01) \\
VCL~\citep{nguyen2017variational}    & 51.8 (\textpm 1.54) & 0.44 (\textpm 0.01) & 48.2 (\textpm 0.99) & 0.50 (\textpm 0.01) \\
\midrule
VCL-Random~\citep{nguyen2017variational} & 52.3 (\textpm 0.66) & 0.43 (\textpm 0.01) & 54.4 (\textpm 1.44) & 0.44 (\textpm 0.01) \\
AGEM~\citep{chaudhry2019agem}            & 62.1 (\textpm 1.39) & 0.21 (\textpm 0.01) & 50.9 (\textpm 0.92) & 0.40 (\textpm 0.01) \\
MER~\citep{riemer2018learning}           & 69.9 (\textpm 0.40) & 0.14 (\textpm 0.01) & 66.0 (\textpm 2.04) & 0.23 (\textpm 0.01) \\
ER-Ring~\citep{chaudhry2019er}           & 70.2 (\textpm 0.56) & 0.12 (\textpm 0.01) & 65.9 (\textpm 0.41) & 0.24 (\textpm 0.01) \\
MIR~\citep{mir_aljundi}                  & 71.1 (\textpm 0.41) & 0.11 (\textpm 0.01) & - & - \\
\textbf{\ours{} (ours)} & \textbf{73.6} (\textpm 0.31) & \textbf{0.09} (\textpm 0.01) & \textbf{68.4} (\textpm 0.72) & \textbf{0.21} (\textpm 0.01) \\
\midrule
Clone-and-finetune   & 81.4 (\textpm 0.35) & 0.0 & 87.5 (\textpm 0.11) & 0.0 \\
Multitask & 83.0 & 0.0 & 83.3 & 0.0 \\
\bottomrule
\end{tabular}

\vskip 0.5cm

\begin{tabular}{lcccc}
\toprule
\multicolumn{1}{l}{\textbf{Method}}  &\multicolumn{2}{c}{\textbf{Split CIFAR}}  &\multicolumn{2}{c}{\textbf{Split miniImageNet}} \\
\midrule
& Accuracy & Forgetting & Accuracy & Forgetting \\
\midrule
Finetune                             & 42.9 (\textpm 2.07) & 0.25 (\textpm 0.03) & 34.7 (\textpm 2.69) & 0.26 (\textpm 0.03) \\
EWC~\citep{Kirkpatrick2016EWC}       & 42.4 (\textpm 3.02) & 0.26 (\textpm 0.02) & 37.7 (\textpm 3.29) & 0.21 (\textpm 0.03) \\
\midrule
ICARL~\citep{Rebuffi16icarl}         & 46.4 (\textpm 1.21) & 0.16 (\textpm 0.01) & - & - \\
AGEM~\citep{chaudhry2019agem}        & 54.9 (\textpm 2.92) & 0.14 (\textpm 0.03) & 48.2 (\textpm 2.49) & 0.13 (\textpm 0.02)\\
MER~\citep{riemer2018learning}       & 49.7 (\textpm 2.97) & 0.19 (\textpm 0.03) & 45.5 (\textpm 1.49) & 0.15 (\textpm 0.01) \\
ER-Ring~\citep{chaudhry2019er}       & 56.2 (\textpm 1.93) & 0.13 (\textpm 0.01) & 49.0 (\textpm 2.61) & 0.12 (\textpm 0.02)\\
MIR~\citep{mir_aljundi}              & 57.1 (\textpm 1.81) & 0.12 (\textpm 0.01) & 49.3 (\textpm 2.15) & 0.12 (\textpm 0.01)\\
\textbf{\ours{} (ours)} & \textbf{60.4} (\textpm 0.54) & \textbf{0.10} (\textpm 0.01) & \textbf{51.6} (\textpm 2.02) & \textbf{0.10} (\textpm 0.01) \\
\midrule
Clone-and-finetune                   & 60.3 (\textpm 0.55) & 0.0  & 50.3 (\textpm 1.00) & 0.0 \\
Multitask                            & 68.3 & 0.0 & 63.5 & 0.0 \\
\bottomrule
\end{tabular}
\label{table:results}
\end{sc}
\end{small}
\end{center}
\end{table*}

\begin{figure*}[t]
		\def \SUBWIDTH {0.24\linewidth}
		\def \FIGSCALE {0.45}
		\def \HORZSPACE {-0.25em}
        \begin{subfigure}{\SUBWIDTH}
        \begin{center}
            \includegraphics[scale=\FIGSCALE]{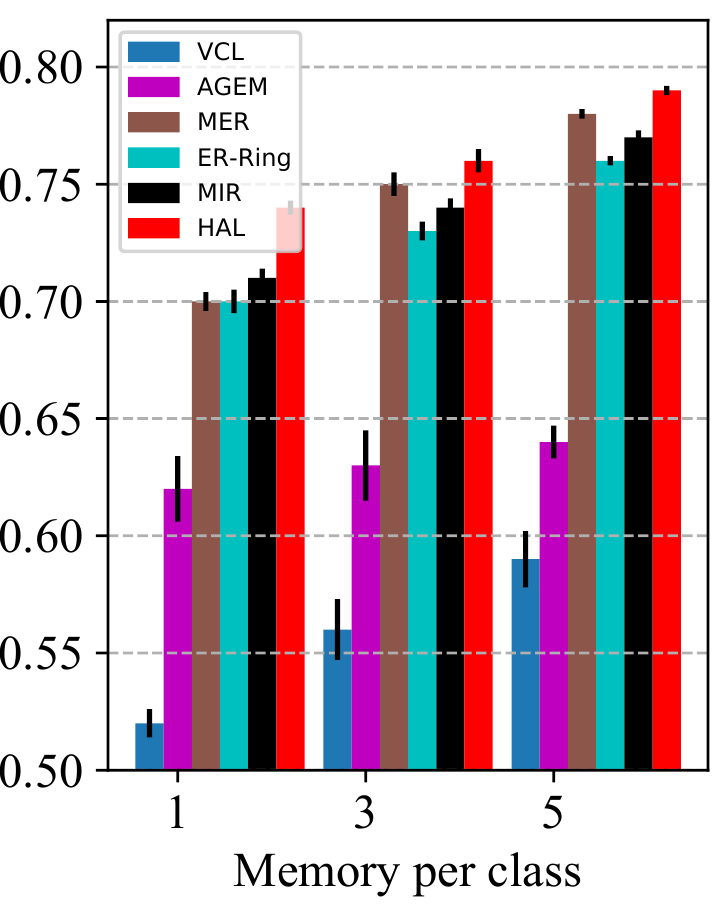}
        \end{center}
        \caption{\emph{\small Permuted MNIST}}
        \end{subfigure}\hspace{\HORZSPACE}
        \begin{subfigure}{\SUBWIDTH}
        \begin{center}
                \includegraphics[scale=\FIGSCALE]{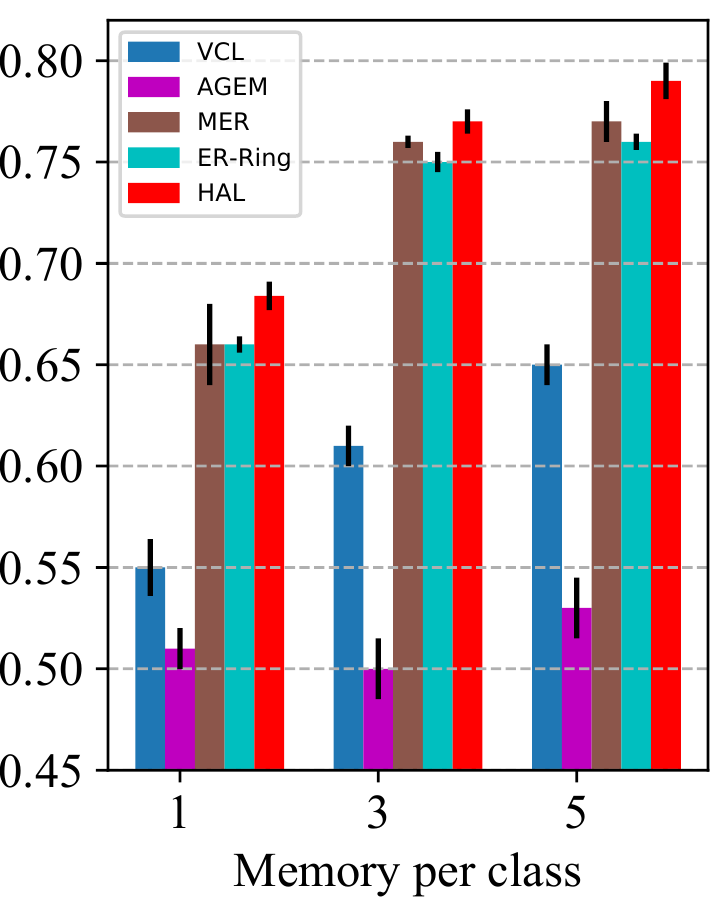}
        \end{center}
        \caption{\emph{\small Rotated MNIST}}
        \end{subfigure}\hspace{\HORZSPACE}
        \begin{subfigure}{\SUBWIDTH}
        \begin{center}
                \includegraphics[scale=\FIGSCALE]{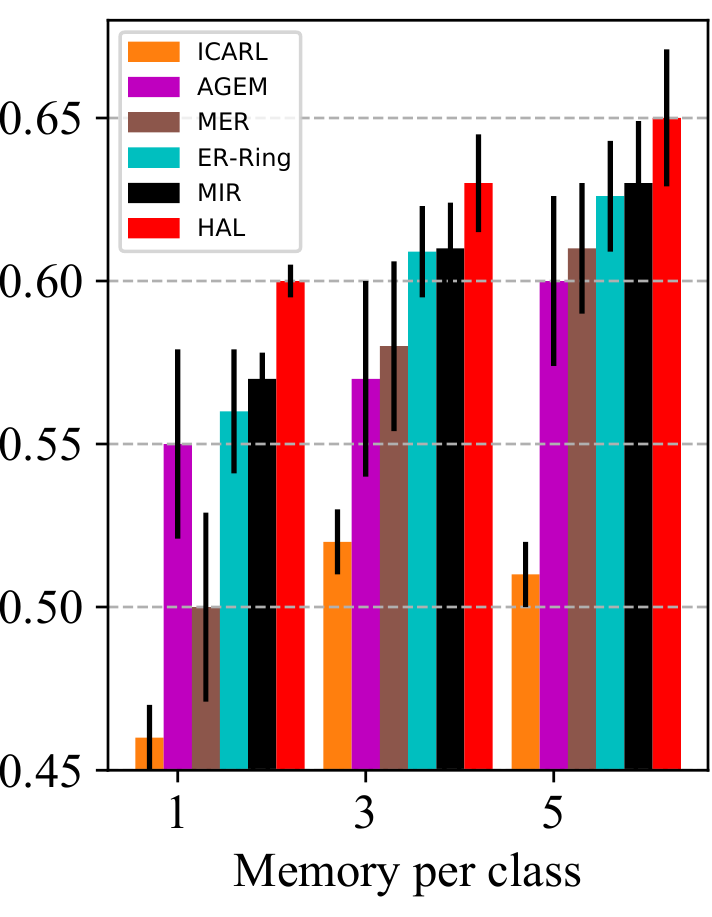}
        \end{center}
        \caption{\emph{\small Split CIFAR}}
        \end{subfigure}\hspace{\HORZSPACE}
        \begin{subfigure}{\SUBWIDTH}
        \begin{center}
            \includegraphics[scale=\FIGSCALE]{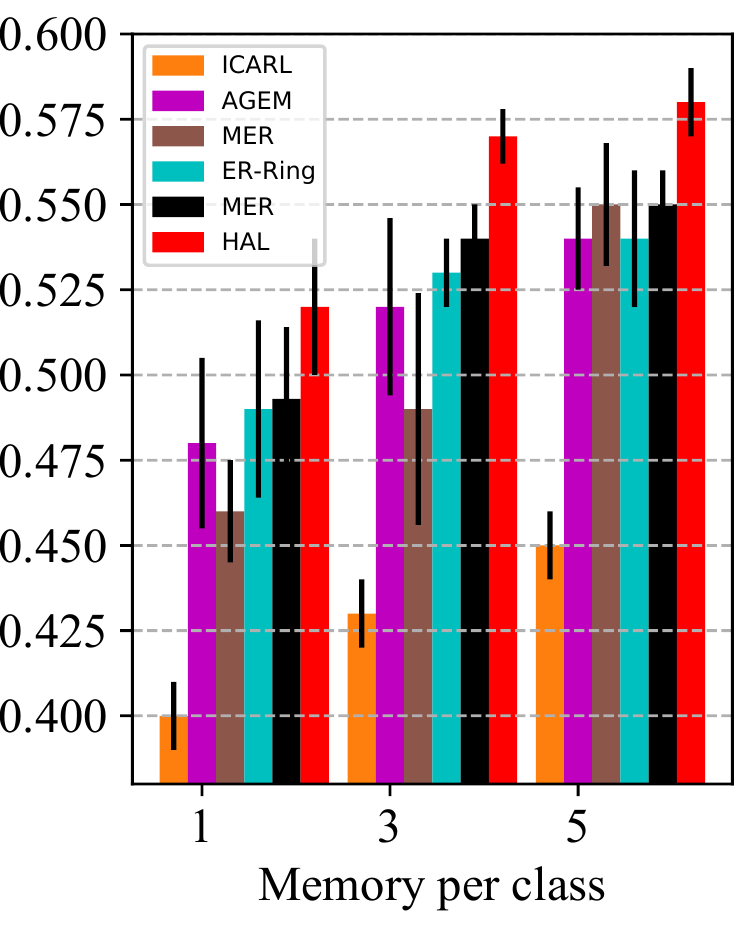}
        \end{center}
        \caption{\emph{\small Split miniImageNet}}
        \end{subfigure}\hspace{0.9em}
    \caption{\emph{Accuracy~(\eqref{eq:accuracy}) results for different episodic memory sizes.}}
\label{fig:big_memories}
\end{figure*}

Table~\ref{table:results} summarizes the main results of our experiments when the episodic memory of only one example per class per task is used.
First, our proposed \emph{\ours{}} is the method achieving maximum Accuracy~(\eqref{eq:accuracy}) and minimal Forgetting~(\eqref{eq:forgetting}) for all benchmarks.
This does not include Oracle baselines \emph{Multitask} (which has access to all data simultaneously) and \emph{Clone-and-finetune} (which trains a separate model per task).
Second, the relative gains from \emph{ER-Ring} to \emph{\ours{}} are substantial, confirming that the anchoring objective~(\eqref{eq:main_update}) allows experience-replay methods to generalize better with the same amount of episodic memory. 

Third, regularization based approaches, such as EWC~\citep{Kirkpatrick2016EWC} and VCL~\citep{nguyen2017variational}, suffer under the single epoch setup. As noted by~\citet{chaudhry2019agem}, EWC requires multiple passes over the samples of each task to perform well. The poor performance of VCL is attributed to the noisy posterior estimation in the single pass setup. Note that approaches making use of memory (MER, ER, MIR, and \ours{}) work substantially better in this setup. 

Fourth, ICARL~\citep{Rebuffi16icarl}, another method making use of episodic memory, performs poorly in our setup. From Table~\ref{table:results}, it can be argued that direct training on a very small episodic memory, as done in experience replay, allows the method to generalize better compared to when the same memory is used indirectly in the knowledge distillation loss~\citep{hinton2015distilling} as done in ICARL.  

Fig~\ref{fig:big_memories} shows the accuracy (\eqref{eq:accuracy}) of methods employing episodic memory when the size of memory is increased. We use $1$ to $5$ examples per class per task, resulting in a total memory size from $200$ to $1000$ for MNIST experiments, and from $85$ to $425$ for CIFAR and ImageNet experiments.
The corresponding numbers for Forgetting are given in Appendix~\ref{sec:more_results}.
\ours{} consistently improves on \emph{ER-Ring} and other baselines.

Figure~\ref{fig:mnist_time} in Appendix~\ref{sec:more_results} provides the training time of the continual learning baselines on MNIST benchmarks. 
Although \ours{} adds an overhead on top of the experience replay baseline, it is substantially faster than MER ---another approach that makes use of nested optimization to reduce forgetting.
However, \ours{} requires extra memory to store task anchors that, as we will show next, are more effective than additional data samples one can store for experience replay. 
Overall, we conclude that \ours{} provides the best trade-off in terms of efficiency and performance.

\subsection{Ablation Study} \label{sec:abal_studies}
We now turn our attention towards two questions; ($1$) whether for the same episodic memory size in bytes \ours{} improves over the experience replay baseline, ($2$) whether fine-tuning on the replay buffer is a good approximation of forgetting when the learner is updated on future tasks.   

\begin{figure*}[t]
	\begin{minipage}[t]{0.45\textwidth}
        \begin{center}
                \includegraphics[scale=0.40]{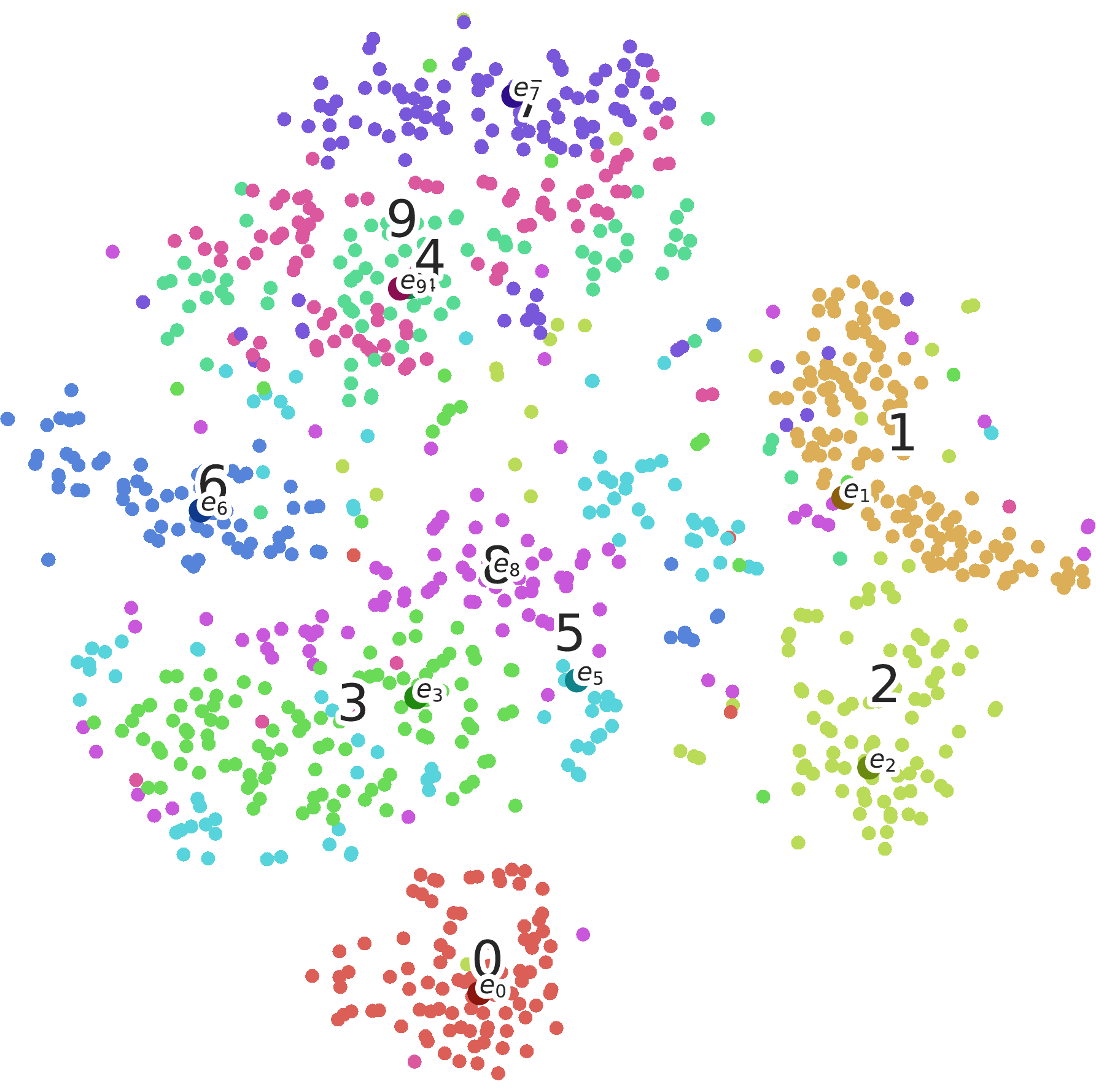}\\
                 {\small Image Space}
        \end{center}
        \end{minipage}\hfill
        \begin{minipage}[t]{0.55\textwidth}
        \begin{center}
                \includegraphics[scale=0.40]{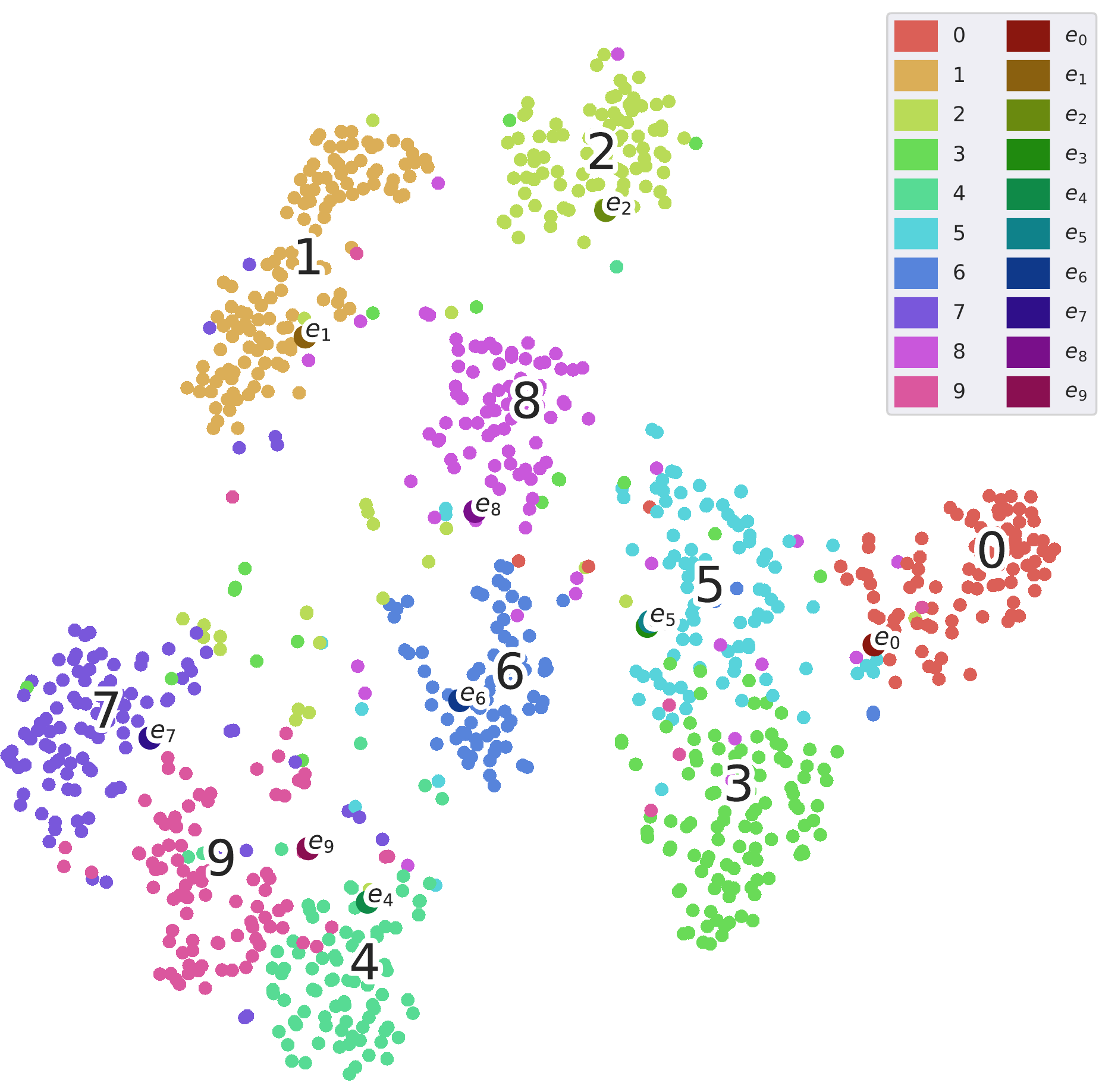}\\
                 {\small Feature Space}
        \end{center}
        \end{minipage}%
\caption{t-SNE visualization of images and anchors (HAL) in the image space (\emph{left}) and the feature space (\emph{right}) on Permuted MNIST benchmark for a single task. Anchor points are exaggerated in size for the purpose of better visualization. The left plot shows that anchor points lie with in the data cluster of a class, whereas the right plot shows that, in the feature space, anchor points lie close to the edge of the cluster of a class or near decision boundaries.}
\label{fig:mnist_visualize}
\end{figure*}

To answer the first question, let $|\epsmem|$ be the total size of episodic memory for all tasks when one example per class per task is stored in the replay buffer. 
We then run the experience replay with double the size of episodic memory (\emph{i.e.}) storing two examples per class per task instead of one.
The episodic memory size in \ours{}, on the other hand, is kept at $|\epsmem|$.
This effectively makes the size of memory in bytes taken by experience replay and that of \ours{} equal as the latter requires extra memory to store anchors. 
Table~\ref{table:twice_data} summarizes the results of this study. 
For the same memory size in bytes, \ours{} performs better than experience replay when \emph{additional real data samples are stored in the episodic memory.} It is surprising that the anchors learned by \ours{}, initialized from random noise and learned using gradient-based optimization, perform better compared to randomly sampled real data. 
To understand this, in Figure~\ref{fig:mnist_visualize} we visualize \ours{}'s anchors along with the task data in the image and feature space on Permuted MNIST benchmark. 
From the left of the figure, it can be seen that \ours{} anchors lie in the data cluster of a class in the image space, suggesting that the mean embedding loss in~\eqref{eq:anchors_2} effectively regularizes against outliers. 
More interestingly, the figure on the right shows that these anchors lie at or close to the cluster edges in the feature space. 
In other words, the anchor points learned by \ours{} lie close to the classifier decision boundary. 
This can explain their effectiveness compared to the real data samples that can lie anywhere in the data cluster in feature space. 

\begin{table}[t]
\begin{center}
\caption{Comparison of \ours{} with experience replay. ER-Ring and \ours{} use one example per class per task in the episodic memory, whereas ER-Ring-$2|\epsmem|$ uses two examples per class per task in the memory. Averages and standard deviations are computed over five runs using different random seeds.}
\vskip -0.10in
\label{table:twice_data}
\resizebox{\columnwidth}{!}{%
\begin{tabular}{lcccc}
\toprule
\multicolumn{1}{l}{\textbf{Method}} & \multicolumn{2}{c}{\textbf{Permuted MNIST}} &\multicolumn{2}{c}{\textbf{Split CIFAR}} \\
\midrule
& Accuracy & Forgetting & Accuracy & Forgetting \\
\midrule
ER-Ring-$|\epsmem|$                              & 70.2 \textpm (0.56) & 0.12 (\textpm 0.01) & 56.2 (\textpm 1.93) & 0.13 (\textpm 0.01) \\
ER-Ring-$2|\epsmem|$                & 71.9 (\textpm 0.31) & 0.11 (\textpm 0.01) & 58.6 (\textpm 2.68) & 0.12 (\textpm 0.01) \\
\textbf{\ours{}-$|\epsmem|$}           & \textbf{73.6} (\textpm 0.31) & \textbf{0.09} (\textpm 0.01) & \textbf{60.4} (\textpm 0.54) & \textbf{0.10} (\textpm 0.01)\\
\bottomrule
\end{tabular}}
\end{center}
\end{table}

\begin{table}[t]
\begin{center}
\caption{Performance comparison of \ours{} with Oracle where the learner has access to all the future tasks to exactly quantify forgettig of an anchor. Averages and standard deviations are computed over five runs using different random seeds. }
\vskip 0.07in
\label{table:oracle_access}
\resizebox{\columnwidth}{!}{%
\begin{tabular}{lcccc}
\toprule
\multicolumn{1}{l}{\textbf{Anchor type}} & \multicolumn{2}{c}{\textbf{Permuted MNIST}} &\multicolumn{2}{c}{\textbf{Split CIFAR}} \\
\midrule
& Accuracy & Forgetting & Accuracy & Forgetting \\
\midrule
\ours{}                          & 73.6 (\textpm 0.31) & 0.09 (\textpm 0.01) & 60.4 (\textpm 0.54) & 0.10 (\textpm 0.01) \\
Oracle                           & 73.9 (\textpm 0.41) & 0.09 (\textpm 0.01) & 61.1 (\textpm 0.94) & 0.09 (\textpm 0.01) \\
\bottomrule
\end{tabular}}
\end{center}
\end{table}

Finally, to answer the second part, we assume a non-continual setup where at each step the learner has an oracle access to all future tasks. 
After training on task $t$, the learner is fine-tuned on all future tasks and anchor points are subsequently learned by optimizing idealistic~\eqref{eq:anchors_1}. 
The results are reported in Table~\ref{table:oracle_access}. 
It can be seen from the table that the proposed \ours{} performs very close to the noncontinual oracle baseline. 
This suggests that \ours{}'s approximation of forgetting when the learner is updated on future tasks by replaying past data is effective in many existing continual learning benchmarks. 

\section{Related work}
\label{sec:related_work}

In continual learning~\citep{ring1997child}, also called lifelong learning~\citep{thrun1998lifelong}, a learner addresses a {\em sequence} of changing tasks without storing the complete datasets of these tasks.
This contrasts with {\em multitask learning}~\citep{caruana1997multitask}, where the learner assumes simultaneous access to data from all tasks.
The main challenge in continual learning is to avoid catastrophic interference~\citep{mccloskey1989catastrophic,mcclelland1995there,goodfellow2013empirical}, that is, the learner forgetting previously acquired knowledge when learning new tasks.
The state-of-the art methods in continual learning can be categorized into three classes.

First, \emph{regularization approaches} discourage updating parameters important for past tasks~\citep{Kirkpatrick2016EWC,aljundi2017memory,nguyen2017variational,Zenke2017Continual}.
While efficient in terms of memory and computation, these approaches suffer from brittleness due to feature drift for large number of tasks~\citep{titsias2019functional}.
Additionally, these approaches are only effective when the learner can perform multiple passes over each task~\citep{chaudhry2019agem}, a case deemed unrealistic in this work.

Second, \emph{modular approaches} use different parts of the prediction function for each new task~\citep{fernando2017pathnet,aljundi2017expert,rosenbaum2018routing,chang2018,rcl2018,modularmetal2018}.
Modular approaches do not scale to a large number of tasks, as they require searching over the combinatorial space of module architectures.
Another modular approach~\citep{rusu2016progressive,lee2017lifelong} adds new parts to the prediction function as new tasks are learned.
By construction, modular approaches have zero forgetting, but their memory requirements increase with the number of tasks.

Third, \emph{episodic memory approaches} maintain and revisit a small episodic memory of data from past tasks.
In some of these methods~\citep{li2016learning,Rebuffi16icarl}, examples in the episodic memory are replayed and predictions are kept invariant by means of distillation~\citep{hinton2015distilling}.
In other approaches~\citep{lopez2017gradient,chaudhry2019agem,aljundi2019online} the episodic memory is used as an optimization constraint that discourages increases in loss of past tasks.
More recently, several works~\citep{hayes2018memory,riemer2018learning,rolnick18,chaudhry2019er} have shown that directly optimizing the loss of episodic memory, also known as experience replay, is cheaper than constraint-based approaches and improves the prediction performance.
Our contribution in this paper has been to improve \emph{experience replay methods} with task anchors learned in hindsight.

There are other definitions of continual learning, such as the one of task-free continual learning.
The task-free formulation does not consider the notion of tasks, and instead works on undivided data streams~\citep{aljundi2019task,aljundi2019online}.
We have focused on the task-based definition of continual learning and, similar to many recent works~\citep{lopez2017gradient,hayes2018memory,riemer2018learning,chaudhry2019agem}, assumed that only a {\em single pass through the data} was possible.

Finally, our gradient-based learning of anchors bears a similarity to ~\citep{Simonyan14Saliency} and \citep{wang2018dataset}. 
In~\citet{Simonyan14Saliency}, the authors use gradient ascent on class scores to find saliency maps of a classification model. 
Contrary to them, our proposed hindsight learning objective optimizes for the forgetting metric, as reducing it is necessary for continual learning. 
Dataset distillation~\citep{wang2018dataset} proposes to encode the entire dataset in a few synthetic points at a given parameter vector by a gradient-based optimization process. 
Their method requires access to the entire dataset of a task for optimization purposes. 
We, instead, learn anchors in hindsight from the replay buffer of past tasks \emph{after} training is finished for the current task. 
While~\citet{wang2018dataset} aim to replicate the performance of the entire dataset from the synthetic points, we focus on reducing forgetting of an already learned task.

\section{Conclusion} \label{sec:conclusion}

We introduced a bilevel optimization objective, dubbed anchoring, for continual learning.
In our approach, we learned one ``anchor point'' per class per task, where predictions are requested to remain invariant by means of nested optimization.
These anchors are learned using gradient-based optimization and represent points that would maximize the forgetting of the current task throughout the entire learning experience.
We simulate the forgetting that would happen during the learning of future tasks \emph{in hindsight}, that is, by taking temporary gradient steps across a small episodic memory of past tasks.
We call our approach Hindsight Anchor Learning (\ours{}).
As shown in our experiments, anchoring in hindsight complements and improves the performance of continual learning methods based on experience replay, achieving a new state of the art on four standard continual learning benchmarks.

\section*{Acknowledgement}
The authors would like to thank Marc'Aurelio Ranzato for helpful discussions. This work was supported by the ERC grant ERC-2012-AdG 321162-HELIOS, EPSRC grant Seebibyte EP/M013774/1 and EPSRC/MURI grant EP/N019474/1. We would also like to acknowledge the Royal Academy of Engineering and FiveAI. AC is funded by Amazon Research award. 
\bibliography{myBibliography}
\clearpage
\clearpage
\newpage
\onecolumn
\section*{Appendix}
Section~\ref{sec:approx_grad} describes the approximate update performed by the anchoring objective (\eqref{eq:main_update} in the main paper). Section~\ref{sec:more_results} reports more experimental results. Section~\ref{sec:supp_hyperparam} provides the grid considered for hyper-parameters. Section~\ref{sec:detailed_algo} gives the pseudocode for \ours{}. 

\appendix
\section{Approximate Update of Anchoring Objective} \label{sec:approx_grad}

Here we will use a Taylor series expansion to approximate the update performed by the anchoring objective (\eqref{eq:main_update} in the main paper). In particular, we are interested in the regularization part of the anchoring objective that involves a nested update. We refer to this gradient as $g_{anc}$. We follow similar arguments as~\citep{metareptile}.

Let $\theta_0$ be the parameter vector before the temporary update of the anchoring objective (\eqref{eq:main_update}). Moreover, let $\ell_{ce}$ and $\ell_{L2}$ be the cross-entropy and L2 losses, respectively. We use the following definitions:
\begin{align*}
    \overline{g}_0 &= \ell_{ce}'(\theta_0) \quad \quad (\textrm{gradient of cross-entropy loss at initial point on $\mathcal{B} \cup \mathcal{B}_\mathcal{M}$}) \\
    \overline{H}_0 &= \ell_{ce}''(\theta_0) \quad \quad (\textrm{Hessian of cross-entropy loss at initial point on $\mathcal{B} \cup \mathcal{B}_\mathcal{M}$}) \\
    \overline{g}_1 &= \ell_{L2}'(\theta_0) \quad \quad (\textrm{gradient of L2 loss at initial point on anchors}) \\
    \overline{H}_1 &= \ell_{L2}''(\theta_0) \quad \quad (\textrm{gradient of L2 loss at initial point on anchors}) \\
\end{align*}

Let $U_0 = \theta_0 - \alpha \overline{g}_0$ be the operator giving a temporary update in the two-step process of (\eqref{eq:main_update}), and let $\theta_1$ be the temporary update itself \emph{(i.e.)} $\theta_1 := U_0$ (note that $\tilde{\theta}$ is used in the main paper instead of $\theta_1$). The $g_{anc}$ is given by:

\begin{align*}
    g_{anc} &= \frac{\partial }{\partial \theta_0} \ell_{L2}(U_0) \\
    &= U_0' \cdot \ell'_{L2}(\theta_1) \\
    &= \left(I - \alpha \overline{H}_0 \right) \cdot \ell'_{L2}(\theta_1)  \numberthis \label{eq:anch_1},
\end{align*}

where the second step is obtained by using the chain rule. Now, if we calculate the first order Taylor series approximation of $\ell'_{L2}(\theta_1)$,

\begin{align*}
    \ell'_{L2}(\theta_1) &= \ell'_{L2} (\theta_0) + \ell''_{L2} (\theta_0) \cdot (\theta_1  - \theta_0) + O(||\theta_1-\theta_0||^2) \\
    &= \overline{g}_1 + \overline{H}_1 \cdot (\theta_0 - \alpha \overline{g}_0 - \theta_0) + O(\alpha^2) \\
    &= \overline{g}_1 - \alpha\overline{H}_1 \cdot \overline{g}_0 + O(\alpha^2) \numberthis \label{eq:taylor_update},
\end{align*}

where in the second step we substituted the value of $\theta_1$. By putting \eqref{eq:taylor_update} in \eqref{eq:anch_1} and after some simplification we get:

\begin{equation} \label{eq:grad_form}
    g_{anc} = \overline{g}_1 - \alpha (\overline{H}_1 \cdot \overline{g}_0 + \overline{H}_0 \cdot \overline{g}_1) + O(\alpha^2).
\end{equation}

This form is very similar to the second-order MAML gradient formulation, Eq. 25 in \citep{metareptile}. Further simplification of the inner product terms between Hessian and gradient yields the inner product between the gradients $\overline{g_0}$ and $\overline{g_1}$. This shows that similar to MAML~\citep{finn2017model}, Reptile~\citep{metareptile} and MER~\citep{riemer2018learning}, the anchoring objective, as described in \eqref{eq:main_update} of the main paper, maximizes the inner product between the gradients. However, unlike the other meta-learning approaches, in the anchoring objective, these gradients correspond to different loss functions, cross-entropy and L2 losses on data from the current task and episodic memory, and \ours{} anchors, respectively.

\section{More Results} \label{sec:more_results}
\begin{figure*}[!h]
	\begin{minipage}{0.45\textwidth}
        \begin{center}
                \includegraphics[scale=0.45]{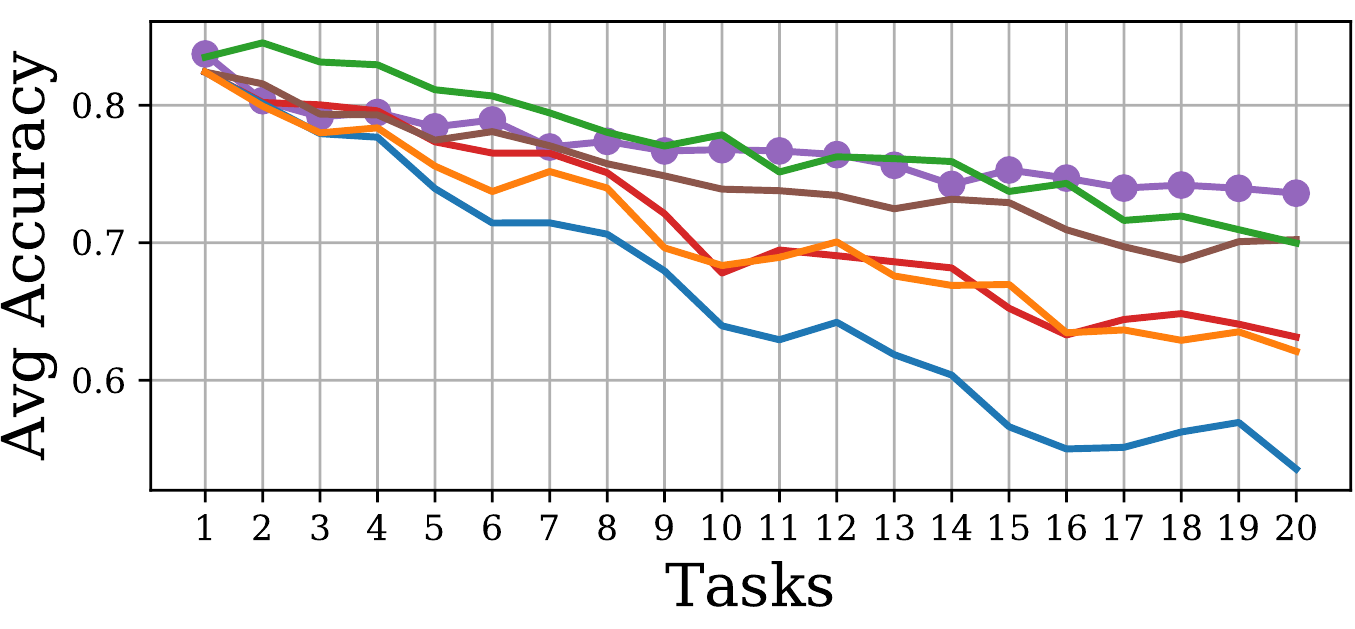}\\
                {\small Permuted MNIST}
        \end{center}
        \end{minipage}\hfill
        \begin{minipage}{0.55\textwidth}
        \begin{center}
                \includegraphics[scale=0.45]{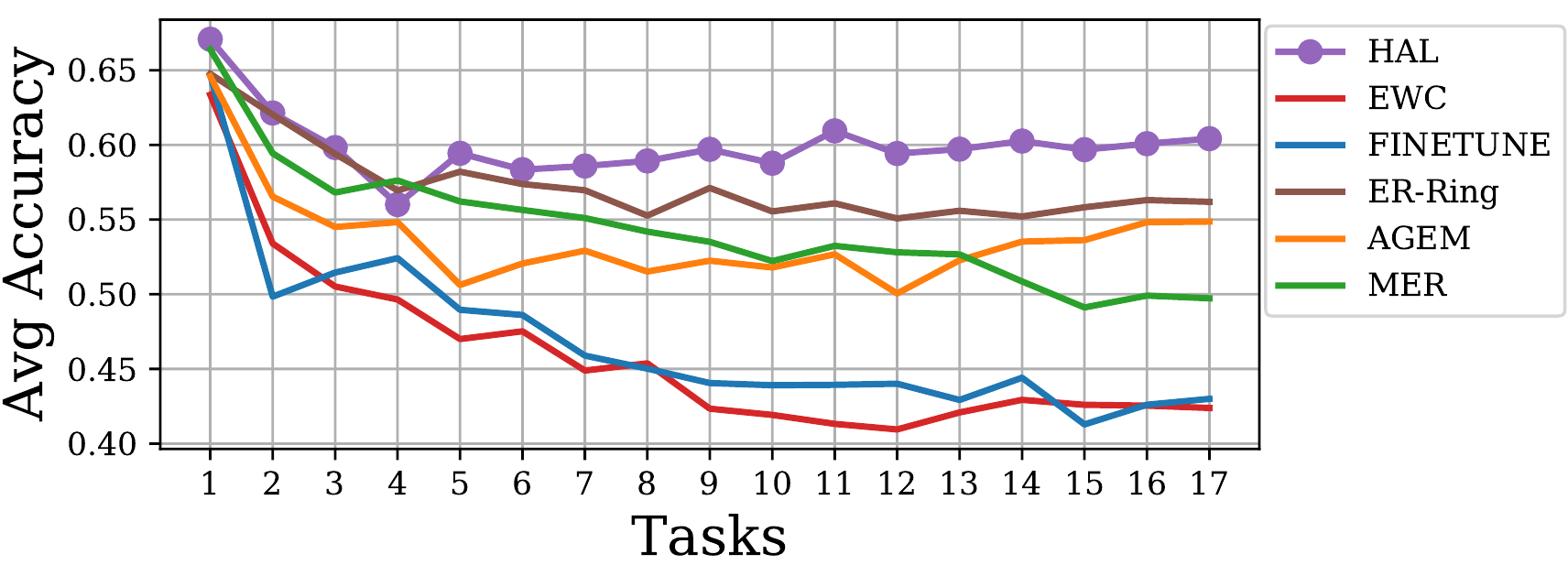} \\
                {\small Split CIFAR}
        \end{center}
        \end{minipage}%
\caption{Evolution of Accuracy~(\eqref{eq:accuracy}) as new tasks are learned. When used, episodic memories contain up to one example per class per task.}
\label{fig:avg_acc_evol}
\end{figure*}

Figure~\ref{fig:avg_acc_evol} shows a more fine-grained analysis of average accuracy as new tasks are learned on Permuted MNIST and Split CIFAR. 
\ours{} preserves the performance of a predictor more effectively than other baselines.

Tables~\ref{table:acc_big_memories} and ~\ref{table:fgt_big_memories} show the accuracy and forgetting of methods employing episodic memory when the size of memory is increased. We use $3$ to $5$ examples per class per task, resulting in a total memory size from $600$ to $1000$ for MNIST experiments, and from $255$ to $425$ for CIFAR and ImageNet experiments.

\begin{table}[!h]
\centering
\caption{Impact of anchor selection, where we compare a randomly chosen data point as an anchor (Real Data Anchor) with our optimized anchor selection (\ours{}).}
\vskip 0.25cm
\label{table:anchor_types}
\begin{tabular}{lcc}
\toprule
\multicolumn{1}{l}{\textbf{Anchor type}} & \multicolumn{2}{c}{\textbf{Split CIFAR}} \\
\midrule
& Accuracy & Forgetting \\
\midrule
Real Data Anchor                         & 58.0 (\textpm 0.15) & 0.12 (\textpm 0.01) \\
\textbf{\ours{} (ours)}                  & 60.4 (\textpm 0.54) & 0.10 (\textpm 0.01)\\
\bottomrule
\end{tabular}
\end{table}

\begin{figure}[!h]
    \begin{center}
                \includegraphics[scale=0.5]{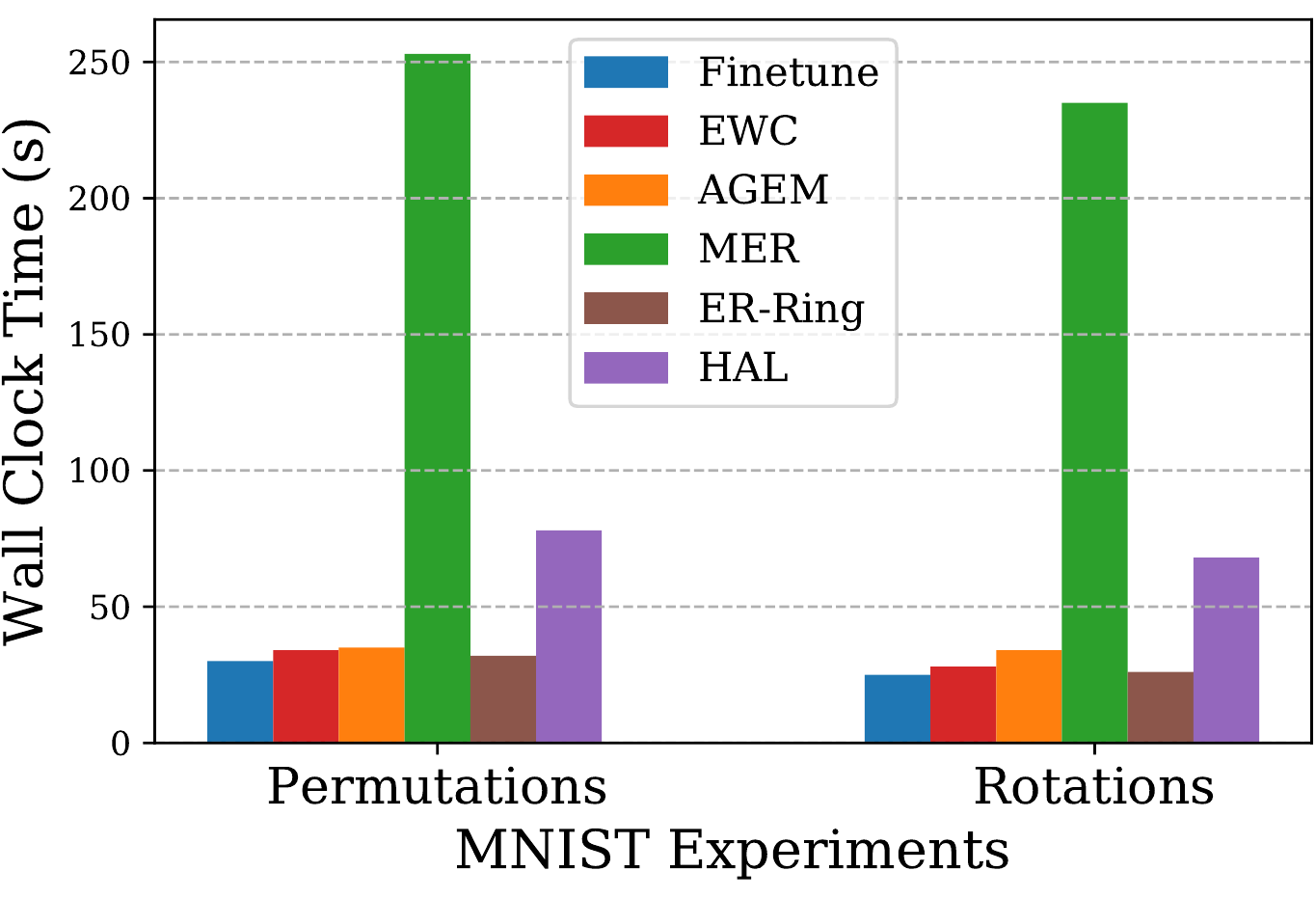}
        \end{center}
    \vspace{-3mm}
    \caption{Training time (s) of MNIST experiments for the entire continual learning experience. MER and \ours{} both use meta-learning objectives to reduce forgetting.}
        \label{fig:mnist_time}
\end{figure}

\begin{table}[!h]
\begin{center}
\begin{small}
\begin{sc}
\caption{Accuracy~(\eqref{eq:accuracy}) results for large ($3$ to $5$ examples per class per task) episodic memory sizes.
Here we only compare methods that use an episodic memory.
Metrics are averaged over five runs using different random seeds.}
\begin{tabular}{lcccc}
\toprule
\multicolumn{1}{l}{\textbf{Method}} & \multicolumn{2}{c}{\textbf{Permuted MNIST}} &\multicolumn{2}{c}{\textbf{Rotated MNIST}} \\
\midrule
& $|\mathcal{M}|=600$ & $|\mathcal{M}|=1000$ & $|\mathcal{M}|=600$ & $|\mathcal{M}|=1000$ \\
\midrule
VCL-Random                           & 55.8 (\textpm 1.29) & 58.5 (\textpm 1.21) & 61.2 (\textpm 0.12) & 64.4 (\textpm 0.16) \\
AGEM                                 & 63.2 (\textpm 1.47) & 64.1 (\textpm 0.74) & 49.9 (\textpm 1.49) & 53.0 (\textpm 1.52) \\
MER                                  & 74.9 (\textpm 0.49) & 78.3 (\textpm 0.19) & 76.5 (\textpm 0.30) & 77.3 (\textpm 1.13) \\
ER-Ring                              & 73.5 (\textpm 0.43) & 75.8 (\textpm 0.24) & 74.7 (\textpm 0.56) & 76.5 (\textpm 0.48) \\
\textbf{\ours{} (ours)} & \textbf{76.2} (\textpm 0.52) & \textbf{78.4} (\textpm 0.27) & \textbf{77.0}  (\textpm 0.66) & \textbf{78.7} (\textpm 0.97) \\
\bottomrule
\end{tabular}

\vskip 0.5cm

\begin{tabular}{lcccc}
\toprule
\multicolumn{1}{l}{\textbf{Method}} &\multicolumn{2}{c}{\textbf{Split CIFAR}} &\multicolumn{2}{c}{\textbf{Split miniImageNet}} \\
\midrule
& $|\mathcal{M}|=255$ & $|\mathcal{M}|=425$ & $|\mathcal{M}|=255$ & $|\mathcal{M}|=425$ \\
\midrule
ICARL                                & 51.7 (\textpm 1.41) & 51.2 (\textpm 1.32) & - & - \\
AGEM                                 & 56.9 (\textpm 3.45) & 59.9 (\textpm 2.64) & 51.6 (\textpm 2.69) & 54.3 (\textpm 1.56) \\
MER                                  & 57.7 (\textpm 2.59) & 60.6 (\textpm 2.09) & 49.4 (\textpm 3.43) & 54.8 (\textpm 1.79) \\
ER-Ring                              & 60.9 (\textpm 1.44) & 62.6 (\textpm 1.77) & 53.5 (\textpm 1.42) & 54.2 (\textpm 3.23)\\
\textbf{\ours{} (ours)} & \textbf{62.9} (\textpm 1.49) & \textbf{64.4} (\textpm 2.15) & \textbf{56.5} (\textpm 0.87) & \textbf{57.2} (\textpm 1.54)\\
\bottomrule
\end{tabular}
\label{table:acc_big_memories}
\end{sc}
\end{small}
\end{center}
\end{table}

\begin{table}[!h]
\begin{center}
\begin{small}
\begin{sc}
\caption{Forgetting~(\eqref{eq:forgetting}) results for large ($3$ to $5$ examples per class per task) episodic memory sizes.
Here we only compare methods that use an episodic memory.
Averages and standard deviations are computed over five runs using different random seeds.}

\begin{tabular}{lcccc}
\toprule
\multicolumn{1}{l}{\textbf{Method}} & \multicolumn{2}{c}{\textbf{Permuted MNIST}} &\multicolumn{2}{c}{\textbf{Rotated MNIST}} \\
\midrule
& $|\mathcal{M}|=600$ & $|\mathcal{M}|=1000$ & $|\mathcal{M}|=600$ & $|\mathcal{M}|=1000$ \\
\midrule
VCL-Random                           & 0.39 (\textpm 0.01) &   0.36 (\textpm 0.01) & 0.37 (\textpm 0.01) & 0.33 (\textpm 0.01) \\
AGEM                                 &  0.20 (\textpm 0.01) &  0.19 (\textpm 0.01) &  0.41 (\textpm 0.01) &  0.38 (\textpm 0.01) \\
MER                                  &  0.14 (\textpm 0.01) &  0.09 (\textpm 0.01) &  \textbf{0.12} (\textpm 0.01) &  \textbf{0.11} (\textpm 0.01) \\
ER-Ring                              &  0.09 (\textpm 0.01) &  0.07 (\textpm 0.01) &  0.15 (\textpm 0.01) &  0.13 (\textpm 0.01) \\
\textbf{\ours{} (ours)}              &  \textbf{0.07} (\textpm 0.01) &  \textbf{0.05} (\textpm 0.01) &  \textbf{0.12} (\textpm 0.01) &  \textbf{0.11} (\textpm 0.01) \\
\bottomrule
\end{tabular}

\vskip 0.5cm

\begin{tabular}{lcccc}
\toprule
\multicolumn{1}{l}{\textbf{Method}} &\multicolumn{2}{c}{\textbf{Split CIFAR}} &\multicolumn{2}{c}{\textbf{Split miniImageNet}} \\
\midrule
& $|\mathcal{M}|=255$ & $|\mathcal{M}|=425$ & $|\mathcal{M}|=255$ & $|\mathcal{M}|=425$ \\
\midrule
ICARL                                &  0.13 (\textpm 0.02) & 0.13 (\textpm 0.02) & - & - \\
AGEM                                 &  0.13 (\textpm 0.03) & 0.10 (\textpm 0.02) & 0.10 (\textpm 0.02) & 0.08 (\textpm 0.01) \\
MER                                  &  0.11 (\textpm 0.01) & 0.09 (\textpm 0.02) & 0.12 (\textpm 0.02) & 0.07 (\textpm 0.01) \\
ER-Ring                              &  0.09 (\textpm 0.01) & \textbf{0.06} (\textpm 0.01) & 0.07 (\textpm 0.02) & 0.08 (\textpm 0.02) \\
\textbf{\ours{} (ours)}              &  \textbf{0.08} (\textpm 0.01) & \textbf{0.06} (\textpm 0.01) & \textbf{0.06} (\textpm 0.01) & \textbf{0.06} (\textpm 0.01) \\
\bottomrule
\end{tabular}
\label{table:fgt_big_memories}
\end{sc}
\end{small}
\end{center}
\end{table}

\section{Hyper-parameter Selection}
\label{sec:supp_hyperparam}

In this section, we report the hyper-parameters grid considered for experiments.
The best values for different benchmarks are given in parentheses.

\begin{itemize}
    \item Multitask
        \begin{itemize}
            \item \texttt{learning rate: [0.003, 0.01, 0.03 (CIFAR, miniImageNet), 0.1 (MNIST perm, rot), 0.3, 1.0]}
        \end{itemize}
        
    \item {Clone-and-finetune}
        \begin{itemize}
            \item \texttt{learning rate: [0.003, 0.01, 0.03 (CIFAR, miniImageNet), 0.1 (MNIST perm, rot), 0.3, 1.0]}
        \end{itemize}
        
    \item {Finetune}
        \begin{itemize}
            \item \texttt{learning rate: [0.003, 0.01, 0.03 (CIFAR, miniImageNet), 0.1 (MNIST perm, rot), 0.3, 1.0]}
        \end{itemize}
        
    \item {EWC}
        \begin{itemize}
            \item \texttt{learning rate: [0.003, 0.01, 0.03 (CIFAR, miniImageNet), 0.1 (MNIST perm, rot), 0.3, 1.0]}
            \item \texttt{regularization: [0.1, 1, 10 (MNIST perm, rot, CIFAR, miniImageNet), 100, 1000]}
        \end{itemize}
        
    \item {AGEM}
        \begin{itemize}
            \item \texttt{learning rate: [0.003, 0.01, 0.03 (CIFAR, miniImageNet), 0.1 (MNIST perm, rot), 0.3, 1.0]}
        \end{itemize}
        
    \item {MER}
        \begin{itemize}
            \item \texttt{learning rate: [0.003, 0.01, 0.03 (MNIST, CIFAR, miniImageNet), 0.1, 0.3, 1.0]}
            \item \texttt{within batch meta-learning rate: [0.01, 0.03, 0.1 (MNIST, CIFAR, miniImageNet), 0.3, 1.0]}
            \item \texttt{current batch learning rate multiplier: [1, 2, 5 (CIFAR, miniImageNet), 10 (MNIST)]}
            
        \end{itemize}
        
    \item {ER-Ring}
        \begin{itemize}
            \item \texttt{learning rate: [0.003, 0.01, 0.03 (CIFAR, miniImageNet), 0.1 (MNIST perm, rot), 0.3, 1.0]}
        \end{itemize}
        
    \item {\ours{}}
        \begin{itemize}
            \item \texttt{learning rate: [0.003, 0.01, 0.03 (CIFAR, miniImageNet), 0.1 (MNIST perm, rot), 0.3, 1.0]}
            \item \texttt{regularization ($\lambda$): [0.01, 0.03, 0.1 (MNIST perm, rot), 0.3 (miniImageNet), 1 (CIFAR), 3, 10]}
            \item \texttt{mean embedding strength ($\gamma$): [0.01, 0.03, 0.1 (MNIST perm, rot, CIFAR, miniImageNet), 0.3, 1, 3, 10]}
            \item \texttt{decay rate ($\beta$): 0.5}
            \item \texttt{gradient steps on anchors ($k$): 100}
        \end{itemize}
        
\end{itemize}

\section{Hyperparameter Sensitivity}
\label{sec:supp_hyperparam_sens}
In Table~\ref{table:hparam_sens}, we report the performance of \ours{} against a range of hyperparameters. 
For a given hyperparameter in the table, all the other hyperparameters are set to their optimal values found in Sec~\ref{sec:supp_hyperparam} of the appendix. \ours{} is not sensitive to the choice of hyperparameters. 

\begin{table}[!h]
\begin{center}
\begin{small}
\begin{sc}
\caption{Average Accuracy of \ours{} on different values of hyperparameters.
For a given hyperparameter in the table, all the other hyperparameters are set to their optimal values found in Sec~\ref{sec:supp_hyperparam} of the appendix.} 
\begin{tabular}{lcccccc} 
 \toprule
 Dataset & $\lambda$ & Acc           & $\gamma$ & Acc          & $\beta$ & Acc \\
 \midrule
 \multirow{3}{*}{Permuted MNIST}    & 0.01      & 72.8 \textpm (0.52)   & 0.01  & 73.1 \textpm (0.20)     & 0.1  & 72.5 \textpm (0.95) \\
                                    & 0.1       & 73.6 \textpm (0.31)   & 0.1   & 73.6 \textpm (0.31)     & 0.5  & 73.6 \textpm (0.31) \\
                                    & 1.0       & 73.2 \textpm (0.85)   & 1.0   & 73.4 \textpm (0.41)     & 0.9  & 72.9 \textpm (0.39) \\
\midrule
\multirow{3}{*}{Split CIFAR100}     & 0.01      & 58.5 \textpm (1.25)   & 0.01  & 59.8 \textpm (0.65)     & 0.1  & 58.7 \textpm (1.17) \\
                                    & 0.1       & 59.2 \textpm (0.91)   & 0.1   & 60.4 \textpm (0.54)     & 0.5  & 60.4 \textpm (0.54) \\
                                    & 1.0       & 60.4 \textpm (0.54)   & 1.0   & 60.2 \textpm (1.21)     & 0.9  & 59.6 \textpm (1.05) \\
\bottomrule
\end{tabular}
\label{table:hparam_sens}
\end{sc}
\end{small}
\end{center}
\end{table}

\section{\ours{} Algorithm} \label{sec:detailed_algo}
Algorithm~\ref{alg:detailed_algo} provides a pseudocode for \ours{}. 

\begin{algorithm}
\caption{Training of \ours{} on sequential data $\dataset = \{\dataset_1, \cdots, \dataset_T\}$, with total replay buffer size `$\mbox{mem\_sz}$', learning rate `$\alpha$', regularization strength `$\lambda$', mean embedding decay `$\beta$', mean embedding strength `$\eta$'.} 
\footnotesize
\begin{algorithmic}[1]
    \Procedure{\ours{}}{$\dataset, \mbox{mem\_sz}, \alpha, \lambda, \beta$}
    \State $\epsmem \gets \{\}*\mbox{mem\_sz}$ 
    \State $\{\anc_1, \cdots, \anc_T\} \gets \{\}$
    \For{$t \in \{1, \cdots, T\}$} 
        \State $\phi_t \gets \Vec{0}$
        \For{$\mathcal{B} \sim \dataset_t$} \Comment{{\tiny Sample a batch from current task}} 
            \State $\mathcal{B}_{\epsmem} \sim \epsmem$  \Comment{{\tiny Sample a batch from episodic memory}}
            \State $\tilde{\theta} \gets \theta - \alpha \cdot \nabla_\theta \, \ell(\mathcal{B} \cup \mathcal{B}_\mathcal{M})$ \Comment{{\tiny Temporary parameter update}}
            \State $\theta \gets \theta - \alpha \cdot \nabla_\theta \left( \ell(\mathcal{B} \cup \mathcal{B}_\mathcal{M}) + \lambda \cdot \sum_{t' < t} \left(f_\theta(e_{t'}, t') - f_{\tilde{\theta}}(e_{t'}, t')\right)^2\right)$ \Comment{{\tiny Anchoring objective (\eqref{eq:main_update})}}
            \State $\phi_t \gets \beta \cdot \phi_t + (1-\beta) \cdot \phi(\mathcal{B})$ \Comment{{\tiny Running average of mean embedding}}
            \State $\epsmem \gets \mbox{UpdateMemory}(\epsmem, \mathcal{B})$ \Comment{{\tiny Add samples to a ring buffer}}
        \EndFor
        \State $\anc_t, \theta \gets \mbox{GetAnchors}(\epsmem, \theta, \phi_t, \eta)$  \Comment{{\tiny Get anchors for current task}}
    \EndFor
    \State \textbf{return} $\theta, \epsmem$
  \EndProcedure
\end{algorithmic}

\vskip 0.5cm

\begin{algorithmic}[1]
   \Procedure{GetAnchors}{$\epsmem, \theta_t, \phi_t, \gamma$}
        \State $\theta \gets \theta_t$
        \For{$\mathcal{B}_{\epsmem} \sim \epsmem$}
            \State $\theta \gets \theta - \alpha \cdot \nabla_{\theta} \ell(\mathcal{B}_{\epsmem})$ \Comment{{\tiny Finetune $\theta_t$ by taking SGD steps on the episodic memory}}
        \EndFor
        \State $\theta_{\epsmem} \gets \theta$ \Comment{{\tiny Store the updated parameter}}
        \State $\anc_t \gets \mbox{rand()}$ \Comment{{\tiny Initialize the task anchors}}
        \For{${1,\cdots, k}$}
            \State $e_t \gets e_t + \alpha \cdot \nabla_{e_t} \left( \ell(f_{\theta_\mathcal{M}}(e_t, t), y_t) - \ell(f_{\theta_t}(e_t, t), y_t) - \gamma (\phi(e_t) - \phi_t)^2 \right)$ \Comment{{\tiny Maximize forgetting (\eqref{eq:anchors_2})}}
        \EndFor
        \State \textbf{return} $\anc_t, \theta_t$
  \EndProcedure
\end{algorithmic}
\label{alg:detailed_algo}
\end{algorithm}

\SKIP{
\begin{table}[t]
\begin{center}
\begin{small}
\begin{sc}
\caption{Impact of anchor selection, where we compare random example-per-class anchors (Data-), and our optimized anchor selection (\ours{}). Averages and standard deviations are computed over five runs using different random seeds.}
\vskip -0.10in
\label{table:anchor_types}
\resizebox{\columnwidth}{!}{%
\begin{tabular}{lcccc}
\toprule
\multicolumn{1}{l}{\textbf{Anchor type}} & \multicolumn{2}{c}{\textbf{Permuted MNIST}} &\multicolumn{2}{c}{\textbf{Split CIFAR}} \\
\midrule
& Accuracy & Forgetting & Accuracy & Forgetting \\
\midrule
ER-Ring                              & 70.2 \textpm (0.56) & 0.12 (\textpm 0.01) & 56.2 (\textpm 1.93) & 0.13 (\textpm 0.01) \\
ER+Data-Anchors                         & 73.2 (\textpm 0.30) & 0.10 (\textpm 0.01) & 59.0 (\textpm 0.56) & 0.12 (\textpm 0.01) \\
ER+\textbf{\ours{} (ours)}              & \textbf{73.6} (\textpm 0.31) & \textbf{0.09} (\textpm 0.01) & \textbf{60.4} (\textpm 0.54) & \textbf{0.10} (\textpm 0.01)\\
\bottomrule
\end{tabular}}
\end{sc}
\end{small}
\end{center}
\end{table}}
\end{document}